\documentclass[pdflatex,sn-nature]{sn-jnl}

\usepackage{graphicx}%
\usepackage{multirow}%
\usepackage{amsmath,amssymb,amsfonts}%
\usepackage{amsthm}%
\usepackage{mathrsfs}%
\usepackage[title]{appendix}%
\usepackage{xcolor}%
\usepackage{textcomp}%
\usepackage{manyfoot}%
\usepackage{booktabs}%
\usepackage{algorithm}%
\usepackage{algorithmicx}%
\usepackage{algpseudocode}%
\usepackage{listings}%
\usepackage{lineno}%
\usepackage{nameref}

\usepackage{geometry}%
\theoremstyle{thmstyleone}%
\theoremstyle{thmstyletwo}%

\theoremstyle{thmstylethree}%

\newcommand{\framework}{{MS-Emulator}}
\newcommand{\norm}[1]{\left\lVert#1\right\rVert}

\begin{document}

\title[Scaling Musculoskeletal Emulation]{Scaling Whole-Body Human Musculoskeletal Behavior Emulation for Specificity and Diversity}



\author[]{\fnm{Yunyue} \sur{Wei}}
\equalcont{These authors contributed equally to this work.}

\author[]{\fnm{Chenhui} \sur{Zuo}}
\equalcont{These authors contributed equally to this work.}

\author[]{\fnm{Shanning} \sur{Zhuang}}

\author[]{\fnm{Haixin} \sur{Gong}}

\author[]{\fnm{Yaming} \sur{Liu}}

\author*[]{\fnm{Yanan} \sur{Sui*}}\email{ysui@tsinghua.edu.cn}



\affil[]{\orgname{Tsinghua University}}



\abstract{
The embodied learning of human motor control requires whole-body neuro-actuated musculoskeletal dynamics, while the internal muscle-driven processes underlying movement remain inaccessible to direct measurement. Computational modeling offers an alternative, but inverse dynamics methods struggled to resolve redundant control from observed kinematics in the high-dimensional, over-actuated system. Forward imitation approaches based on deep reinforcement learning exhibited inadequate tracking performance due to the curse of dimensionality in both control and reward design. Here we introduce a large-scale parallel musculoskeletal computation framework for biomechanically grounded whole-body motion reproduction. By integrating large-scale parallel GPU simulation with adversarial reward aggregation and value-guided flow exploration, the \framework~framework overcomes key optimization bottlenecks in high-dimensional reinforcement learning for musculoskeletal control, which accurately reproduces a broad repertoire of motions in a whole-body human musculoskeletal system actuated by approximately 700 muscles. It achieved high joint angle accuracy and body position alignment for highly dynamic tasks such as dance, cartwheel, and backflip. The framework was also used to explore the musculoskeletal control solution space, identifying distinct musculoskeletal control policies that converge to nearly identical external kinematic and mechanical measurements. This work establishes a tractable computational route to analyzing the specificity and diversity underlying human embodied control of movement. Project page: \href{https://lnsgroup.cc/research/MS-Emulator}{https://lnsgroup.cc/research/MS-Emulator}.
}


\maketitle

\section*{Introduction}\label{Introduction}

\begin{figure}[th]
\centering
\includegraphics[width=1\textwidth]{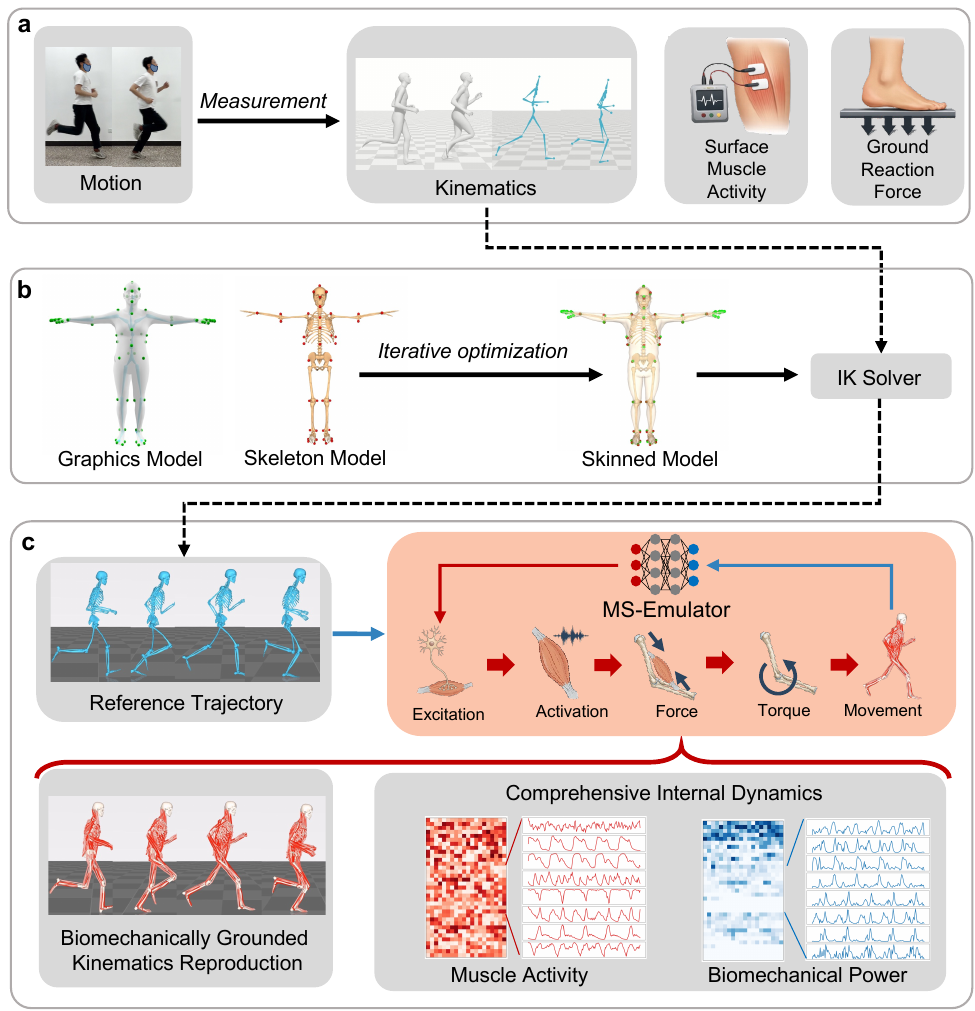}
\caption{\textbf{Dynamics analysis during human motion.} \textbf{a,} In real-world experimental settings, physical measurements are fundamentally restricted to external kinematics, reaction forces (GRF) and surface muscle activity (sEMG). The complex internal musculoskeletal dynamics driving the movement remain unobservable. \textbf{b,} Measured motion is retargeted to the musculoskeletal model by fitting graphical body surface (SMPL-X) to skeletal representations and solving inverse kinematics, yielding a reference trajectory in the model coordinates. \textbf{c,} Using retargeted kinematics as a reference trajectory, \framework~enables biomechanically grounded motion reproduction and reveals plausible internal dynamical solutions under the musculoskeletal model and learning framework.
}
\label{fig_1}
\end{figure}

How internal musculoskeletal dynamics give rise to human movement remains difficult to study directly. A central challenge is to probe control solutions that are both \emph{specific}, in the sense of precisely reproducing a target movement, and \emph{diverse}, in the sense of capturing the multiple feasible internal strategies permitted by the redundant human musculoskeletal system. Here we present a computational framework for inferring plausible internal control solutions from externally measured motion (Fig.~\ref{fig_1}). Human movement emerges from coordinated interactions across more than 200 joints and 600 muscles \cite{wolpert2011principles, uchida2021biomechanics}. Most experiments, however, capture only external kinematics, external forces and a limited set of physiological signals, such as electromyography (EMG). The internal dynamics of movement generation and control therefore remain largely inaccessible (Fig.~\ref{fig_1}a).

Musculoskeletal modeling offers a natural computational route to bridge this measurement gap \cite{damsgaard2006analysis, delp2007opensim, todorov2012mujoco}. In physics-based simulation, musculoskeletal models explicitly represent the forward dynamics linking neural excitation, muscle activation and joint motion \cite{rajagopal2016full}. This representation creates the possibility of investigating internal dynamics from observed movement. Measured motion can be retargeted to a musculoskeletal system through body-surface fitting and inverse kinematics (Fig.~\ref{fig_1}b). The key challenge is then to characterize biologically plausible control from external observations with sufficient specificity for accurate motion reproduction, while preserving the diversity of feasible internal solutions. Whole-body human musculoskeletal systems are highly overactuated, high-dimensional and strongly nonlinear \cite{zuo2024self}. Neuromuscular actuation also introduces intrinsic temporal delays \cite{winters1995improved}. Conventional inverse-dynamics approaches therefore scale poorly to full-body muscle-actuated systems. They often require concurrent external-force measurements, such as ground-reaction forces \cite{anderson2001static, thelen2003generating}, and remain vulnerable to error accumulation during dynamic movement \cite{hicks2015my}.

Recent deep reinforcement learning (DRL) methods provide an alternative route by learning neuromechanical control policies that imitate motion trajectories directly in forward simulation \cite{song2021deep, caggiano2022myosuite, caggiano2023myodex, simos2025kinesis, mimicmjx2025}. Although this perspective avoids explicit inversion, its performance remains insufficient for precise motion analysis. Conventional DRL tracking relies on multi-objective reward functions with manually tuned weights to balance tracking errors across body segments and joints \cite{peng2018deepmimic, lee2019scalable}. Such formulations often compromise both specificity and diversity: they fail to reproduce target motion with sufficient fidelity and restrict exploration in the high-dimensional muscle-actuation space. The full-body muscle-actuation space is extraordinarily large, making the discovery of coordinated control strategies highly inefficient, while realistic muscle-tendon simulation introduces a further computational burden. Consequently, even with dimensionality reduction, curriculum learning and prolonged training, existing methods still struggle to achieve high-fidelity imitation in musculoskeletal systems with realistic full-body anatomical complexity \cite{schumacher2023dep, he2024dynsyn, feng2023musclevae, park2025magnet}.

Here we present an efficient computational pipeline for precise whole-body human motion control and analysis of neuro-actuated musculoskeletal dynamics (Fig.~\ref{fig_1}c). Our \framework~framework combines a full-body musculoskeletal model actuated by 700 independent muscle-tendon units with a high-throughput GPU simulation platform that enables thousands of parallel rollouts on a single device. It further introduces a unified motion-imitation environment that removes manual reward tuning by adaptively aggregating high-dimensional tracking errors, together with a flow-based exploration strategy that improves learning efficiency in the ultra-high-dimensional muscle-actuation space. Together, these components enable efficient learning of muscle-actuated motion from measured kinematics. \framework~reproduces a broad repertoire of human motor behaviors with high kinematic fidelity in the full-body musculoskeletal system. Across motions ranging from walking and running to dancing, cartwheeling and backflipping, the framework maintains low tracking error despite significantly different kinematic and dynamic demands. Emulating a specific running trajectory with an average joint-angle error of $\leq 2^\circ$ requires only 7 hours of training from scratch on a single consumer-grade GPU (NVIDIA 5090).

Leveraging this computational capability, we investigate the solution space of full-body musculoskeletal control during human locomotion. Using distinct regularization schemes during motion tracking, we show that the framework can steer learning towards different regions of the redundant muscle-actuation space while preserving accurate reproduction of the observed movement, supporting both specificity and diversity. It aligns simulated muscle activity with experimental measurements when such agreement is imposed, yet also uncovers multiple distinct internal control strategies enabled by musculoskeletal overactuation. These distinct muscle-control patterns converge to nearly identical kinematic behavior and remain consistent with measured external mechanics. Together, these results show that our pipeline can systematically probe the diversity and structure of internal musculoskeletal dynamics during human motion, providing a computational route to understanding the high-dimensional and redundant control processes that underlie movement beyond observable measurements.

\begin{figure}[ht]
\centering
\includegraphics[width=1\textwidth]{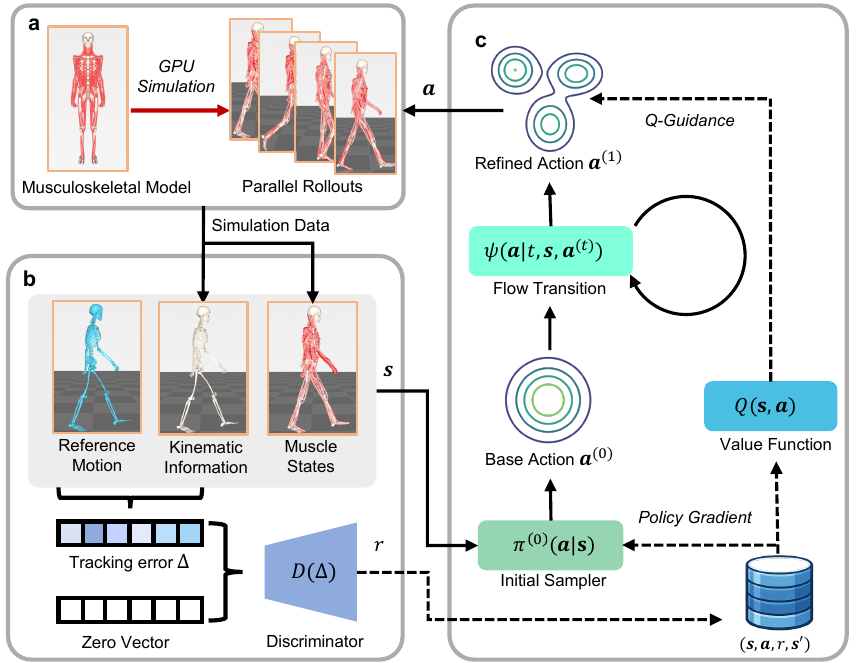}
\caption{\textbf{\framework~enables efficient human musculoskeletal motion reproduction for internal dynamics analysis.} \textbf{a,} A highly detailed, 700-muscle whole-body musculoskeletal model is embedded into GPU simulation to support massively parallel physics rollouts, producing the full system simulation data with neuro-actuated musculoskeletal dynamics. \textbf{b,} Kinematic information and muscle states are extracted from 
simulation data to construct the state $\boldsymbol{s}$ along with reference motion. A discriminator $D(\boldsymbol{\Delta})$, trained to distinguish the tracking-error vector $\boldsymbol{\Delta}$ from a zero vector, provides an adaptive tracking reward $r$. \textbf{c,} Solid arrows denote action generation: an initial sampler $\pi^{(0)}(\boldsymbol{a}|\boldsymbol{s})$ draws a base action from a Gaussian policy, which is then refined by the flow transition $\psi(\boldsymbol{a}|t, \boldsymbol{s}, \boldsymbol{a}^{(t)})$ into the final action $\boldsymbol{a}$ applied to the simulator. Dashed arrows denote learning: replayed transitions $(\boldsymbol{s}, \boldsymbol{a}, r, \boldsymbol{s}')$ are used to optimize the state-value function $Q(\boldsymbol{s}, \boldsymbol{a})$ and the initial sampler by policy gradient, while the flow transition is guided by the learned $Q$-function. The resulting neuro-actuated musculoskeletal dynamics, including neural actuation, muscle force, contact force and temporal dynamics, are recorded for detailed analysis of human motion control.
}
\label{fig_2}
\end{figure}

\section*{Results}\label{Results}

\subsection*{Efficient Framework for Neuro-Actuated Musculoskeletal Motion Control}

To enable high-fidelity motion reproduction, we developed \framework, a large-scale parallel musculoskeletal simulation and control framework that maps generic kinematic trajectories to full-body neuro-actuated musculoskeletal dynamics (Fig.~\ref{fig_2}). Using our framework, reference kinematic trajectories can be reproduced with high accuracy, and comprehensive musculoskeletal movement dynamics can be recorded for detailed analysis. The framework consists of three components: a high-throughput simulation pipeline that embeds a whole-body musculoskeletal model into a GPU-native physics engine to support large-scale parallel rollouts (Fig.~\ref{fig_2}a), a unified imitation environment that provides adaptive reward signals for accurate motion tracking (Fig.~\ref{fig_2}b), and a flow-based RL policy that enables scalable exploration for efficient learning of high-dimensional motion control (Fig.~\ref{fig_2}c). Together, these components make precise whole-body musculoskeletal motion reproduction computationally tractable.

\hfill \break
\noindent\textbf{High-throughput neuro-actuated musculoskeletal simulation}

\noindent
Existing musculoskeletal simulations are typically executed on CPUs \cite{delp2007opensim, caggiano2022myosuite}, which severely constrains the number of parallel rollouts. To address this limitation, we apply a GPU-parallel simulation engine on the MS-Human-700 whole-body musculoskeletal model \cite{zuo2024self} (Fig.~\ref{fig_2}a). By compiling the full musculoskeletal dynamic simulation into GPU kernels via MuJoCo Warp \cite{mujoco_warp} , the engine enables large-scale parallel physics rollouts while retaining detailed internal states, including joint kinematics, muscle activations, muscle forces and joint torques throughout movement. This architecture permits efficient large-scale simulation of complex full-body musculoskeletal dynamics without sacrificing biomechanical fidelity.

\hfill \break
\noindent\textbf{Unified imitation environment with adaptive tracking reward}

\noindent
Conventional reinforcement learning approaches typically rely on multi-objective reward functions that demand laborious, task-specific manual weight tuning to balance errors across various body segments and joints. We bypass this structural limitation by introducing a unified imitation environment powered by an adversarial differential discriminator \cite{zhang2025physics} (Fig.~\ref{fig_2}b). At each simulation step, the environment evaluates a high-dimensional tracking-error vector comprising root translations, root rotations, full-body joint angle discrepancies, and key body segment position deviations. The discriminator learns to distinguish this simulated error from a theoretical zero-error reference, outputting a continuous probability score that serves as the instantaneous reward for the control policy (see \nameref{Methods} for details). As the musculoskeletal agent improves, the discriminator adaptively refines its decision boundary to become increasingly stringent. This adversarial co-evolution establishes an implicit learning curriculum that automatically isolates and penalizes the most challenging unlearned kinematic deviations at any training phase. By providing a smoothly shifting and unified supervision signal, this formulation ensures robust full-body pose alignment across diverse motor behaviors without requiring manual reward engineering.

\hfill \break
\noindent\textbf{Scalable reinforcement learning with flow-based exploration}

\noindent
Existing reinforcement learning algorithms for massively parallel training typically rely on isotropic Gaussian exploration \cite{schulman2017proximal, haarnoja2018soft}. In high-dimensional, over-actuated musculoskeletal systems, this strategy rapidly loses effective coverage of the action space and often induces rigid co-contraction patterns \cite{schumacher2023dep}. To overcome this limitation, we integrate value-guided flow exploration \cite{wei2026scalable} into our on-policy RL pipeline (Fig.~\ref{fig_2}c). The policy is parameterized by an initial action sampler and a flow velocity field. Beyond standard on-policy updates, the velocity field is trained to progressively transport sampled actions towards a target distribution induced by the learned state-action value function. The resulting transported distribution constitutes a valid policy-improvement step, enabling more efficient use of massively parallel sampling. Exploration is performed directly in the original muscle-actuation space. As a result, the method does not impose a low-dimensional control prior, which preserves the full flexibility of the over-actuated musculoskeletal system and allows diverse internal control strategies to emerge under the same motion objective.

\subsection*{Specific and Diverse Whole-Body Musculoskeletal Motion Tracking}\label{subsec:tracking}

\begin{figure}[H]
\centering
\includegraphics[width=1\textwidth]{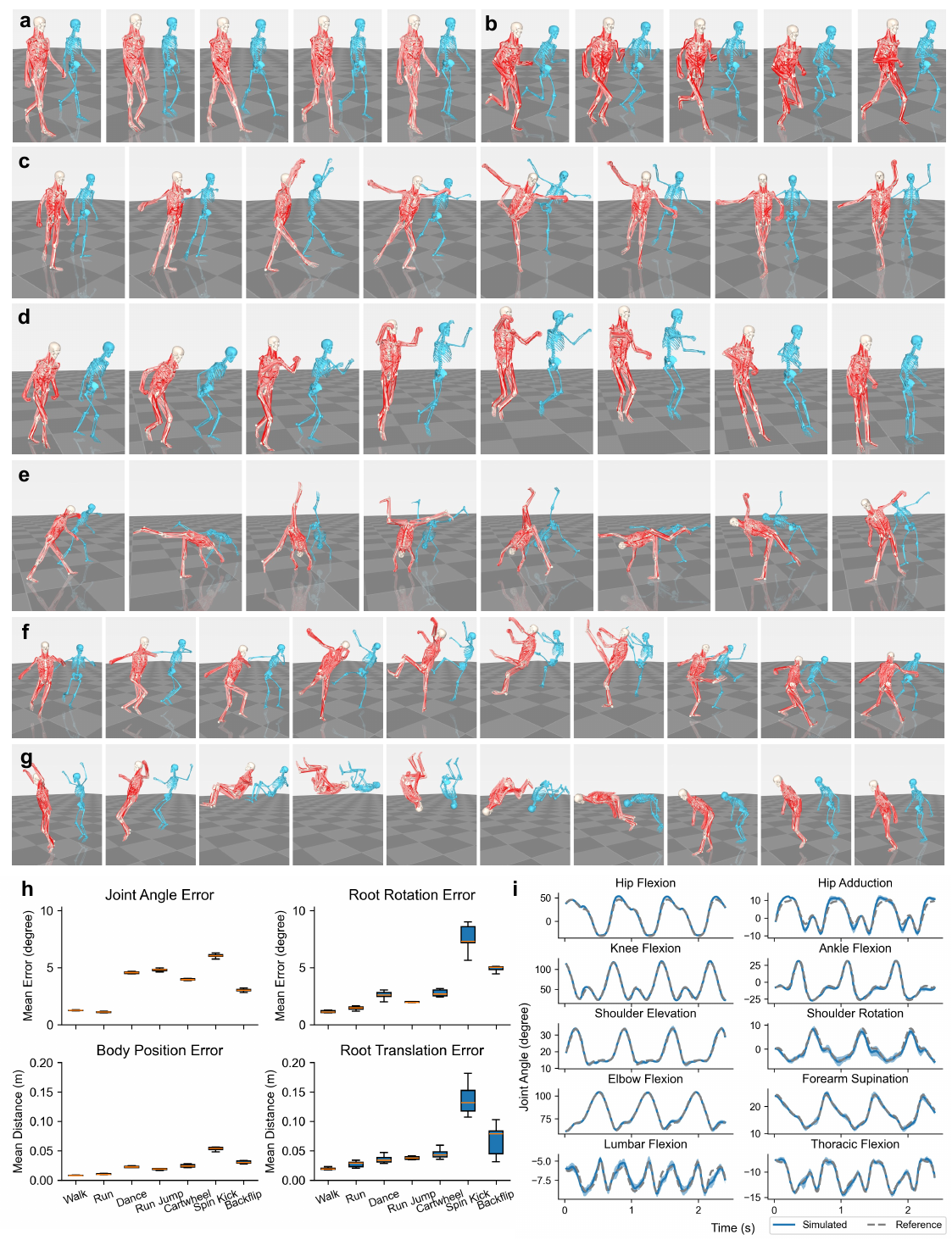}
\caption{\textbf{High-fidelity reproduction of diverse human motor skills.} \textbf{a-g,} Filmstrips demonstrating the simulated full-body musculoskeletal agent (red, rendering muscle activations) tracking a highly diverse repertoire of reference kinematics (blue skeleton). The learned behaviors span cyclic locomotion, including walking (\textbf{a}) and running (\textbf{b}), as well as more agile and dynamically challenging skills such as dance (\textbf{c}), run jump (\textbf{d}), cartwheel (\textbf{e}), spin kick (\textbf{f}) and backflip (\textbf{g}). 
\textbf{h,} Quantitative evaluation of tracking accuracy across tasks. Box plots show converged mean errors for joint angles, root rotation, body position and root translation.
\textbf{i,} Representative time-series comparison of major joint angles during running cycles, illustrating tight temporal and spatial alignment between the simulated closed-loop control (blue solid lines) and the reference kinematic trajectories (grey dashed lines). 
Shaded regions and error bars denote one standard deviation (n=10).}
\label{fig_3}
\end{figure}

To assess the behavioral versatility and kinematic fidelity of the simulated musculoskeletal agent, we used \framework~to reproduce a broad repertoire of human motor skills from reference trajectories derived from the Gait120 and AMASS datasets \cite{boo2025comprehensive, mahmood2019amass}. The framework successfully learned high-dimensional control policies for behaviors ranging from steady cyclic locomotion, such as walking and running, to agile and highly dynamic maneuvers, including dancing, jumping, kicking, cartwheeling and backflipping (Fig.~\ref{fig_3}a--g, Supplementary Videos). In each motion sequence, the simulated musculoskeletal agent closely follows the reference trajectory across motions with very different dynamical and coordination demands. Additional examples, including vertical jump and kick demonstrations, are provided in the Supplementary Videos.

Quantitative evaluation showed consistently accurate tracking across this diverse skill set (Fig.~\ref{fig_3}h). For steady-state locomotion, such as walking, the agent achieved mean joint-angle errors of $\leq 2^\circ$. Even when averaged across all highly dynamic and complex tasks, such as cartwheels, spin kicks and backflips, the framework maintained mean joint-angle errors of $\leq 7^\circ$, together with mean body-position errors of approximately 6~cm. Achieving this level of accuracy across behaviors with markedly different kinematic and dynamic demands, in a system actuated by approximately 700 muscles, demonstrates the specificity of motion reproduction supported by the framework. The successful reproduction of many qualitatively distinct tasks demonstrates the breadth of the motion repertoire addressed here. We analyze solution-space diversity, namely multiple internal control solutions for the same movement, separately in Fig.~\ref{fig_5}.

Temporal analysis of major joint kinematics further confirmed the stability and precision of the learned muscle-actuated policies. As shown in Fig.~\ref{fig_3}i, the simulated trajectories closely matched the phase-dependent evolution of the reference joint kinematics throughout the movement cycle. Variability across repeated trials remained low, with narrow standard-deviation bands around the mean trajectories, indicating robust convergence to consistent control solutions. This fidelity extended beyond major appendicular joints, such as the hip and knee, to more subtle motions of the torso, including lumbar and thoracic flexion. Together, these results show that \framework~supports accurate and robust tracking of diverse whole-body movements, providing a foundation for scalable simulation of high-dimensional motor behavior and internal musculoskeletal dynamics.

\subsection*{Accelerated Learning via Large-scale Parallelism}\label{subsec:efficiency}

The scalability of \framework~is critical because practical musculoskeletal analysis requires not only accurate motion reproduction, but also training and evaluation across large numbers of trajectories, control policies and analysis conditions. Reconstructing control for a whole-body musculoskeletal system actuated by approximately 700 muscles is computationally demanding, and previous CPU-based pipelines have therefore limited both training speed and experimental scale. By executing simulation, observation processing and policy optimization in a unified GPU-native pipeline, we enabled large-scale parallel rollouts without repeated CPU-GPU data transfer.

\begin{figure}[h]
\centering
\includegraphics[width=1\textwidth]{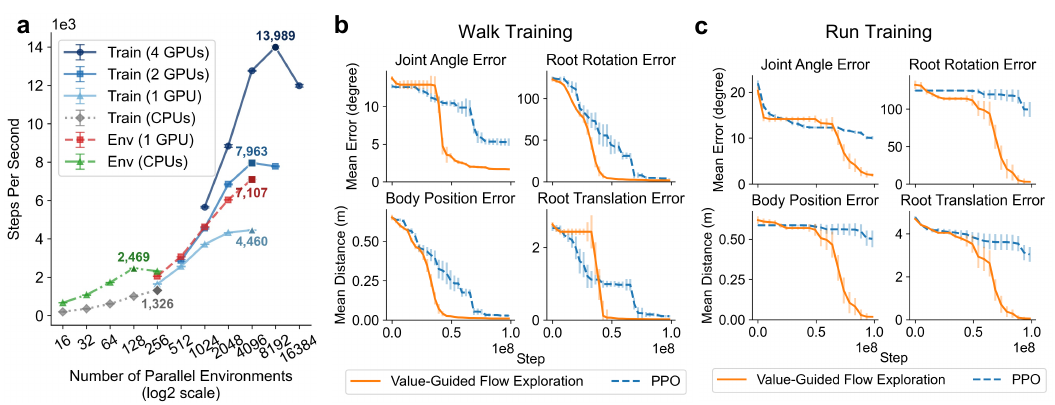}
\caption{\textbf{Large-scale parallelism accelerates musculoskeletal simulation and learning.} \textbf{a,} Throughput benchmark as a function of the number of parallel environments. Green and red curves show environment stepping speed on CPUs and a single GPU under random actions, respectively. Grey and blue curves show end-to-end training throughput for a conventional CPU baseline and our GPU implementation on 1, 2 and 4 GPUs. GPU measurements were obtained on NVIDIA GeForce RTX 5090 GPUs; CPU baselines were measured on the server described in \nameref{Methods}. Throughput increases approximately linearly before saturating. \textbf{b,c,} Best tracking performance during training for walking (\textbf{b}) and running (\textbf{c}), comparing value-guided flow exploration (orange) with PPO (blue dashed). All trainings were performed on a single 5090 GPU. Curves show the mean across three random seeds; vertical error bars denote one standard deviation. Our method converges faster and to lower final errors in both tasks, with the larger advantage observed in the more dynamically demanding running task.}
\label{fig_4}
\end{figure}

We quantified the systems-level gains of \framework~by measuring simulation and training throughput as a function of the number of parallel environments (Fig.~\ref{fig_4}a). The simulator reached a forward-dynamics speed of 7,107 steps per second (SPS) on a single GPU, compared with 2,469 SPS on a cluster with 224 CPU cores. The same advantage persisted during end-to-end policy training. A single NVIDIA GeForce RTX 5090 achieved 4,460 SPS, whereas the conventional CPU baseline \cite{raffin2021stable} reached only 1,326 SPS. Training throughput further increased to 7,963 SPS on 2 GPUs and 13,989 SPS on 4 GPUs. Across configurations, throughput scaled approximately linearly with the number of parallel environments before saturating at hardware limits, indicating that the framework efficiently converts large-scale parallel simulation into usable optimization throughput.

\framework~also improves sample efficiency by incorporating a scalable exploration strategy for policy training. We compared value-guided flow exploration with PPO \cite{schulman2017proximal} on walking and running motion-tracking tasks (Fig.~\ref{fig_4}b,c) using the same simulator, identical environments, and aligned algorithmic hyperparameters. In each evaluation, the agent was initialized at the beginning of the reference trajectory and rolled out to its end, without any early termination. During walk task training, our method reduced all tracking errors more rapidly and converged to lower final errors, with the clearest improvement in joint-angle tracking. The advantage became even larger in running, a more dynamically demanding task requiring tighter full-body coordination. Value-guided flow exploration drove joint, root and body errors close to zero within approximately $10^8$ environment steps, whereas PPO remained at substantially higher error levels throughout training. These results show that the accelerated learning enabled by \framework~arises from two complementary factors: massively parallel GPU simulation, which removes the throughput bottleneck of full-body musculoskeletal control, and directed exploration, which improves optimization in the high-dimensional muscle-actuation space. Together, these advances make training neuro-actuated whole-body motion-tracking controllers feasible within hours on a small number of consumer-grade GPUs and establish the computational basis for large-scale motion analysis.

\subsection*{Analysis of Neuro-Actuated Musculoskeletal Dynamics and Redundancy}\label{subsec:manifold}

Leveraging the rapid training throughput of our framework, we performed a statistical analysis of neuro-actuated musculoskeletal dynamics during walking. Because the human musculoskeletal system is inherently redundant, a single kinematic trajectory can, in principle, be realized by multiple distinct muscle-activation strategies \cite{bernshtein1967co, latash2012bliss}. Walking therefore provides a useful setting for studying how precise behavioral reproduction can coexist with diverse internal control solutions. To probe this redundancy, we trained ensembles of walking-control policies under three objectives (see \nameref{Methods}): (1) imitation-only tracking, (2) imitation with power regularization, which penalizes muscle effort during motion generation, and (3) imitation with EMG regularization, which imposes agreement with experimentally measured sEMG from ten muscle groups as an auxiliary training constraint.

\begin{figure}[H]
\centering
\includegraphics[width=1\textwidth]{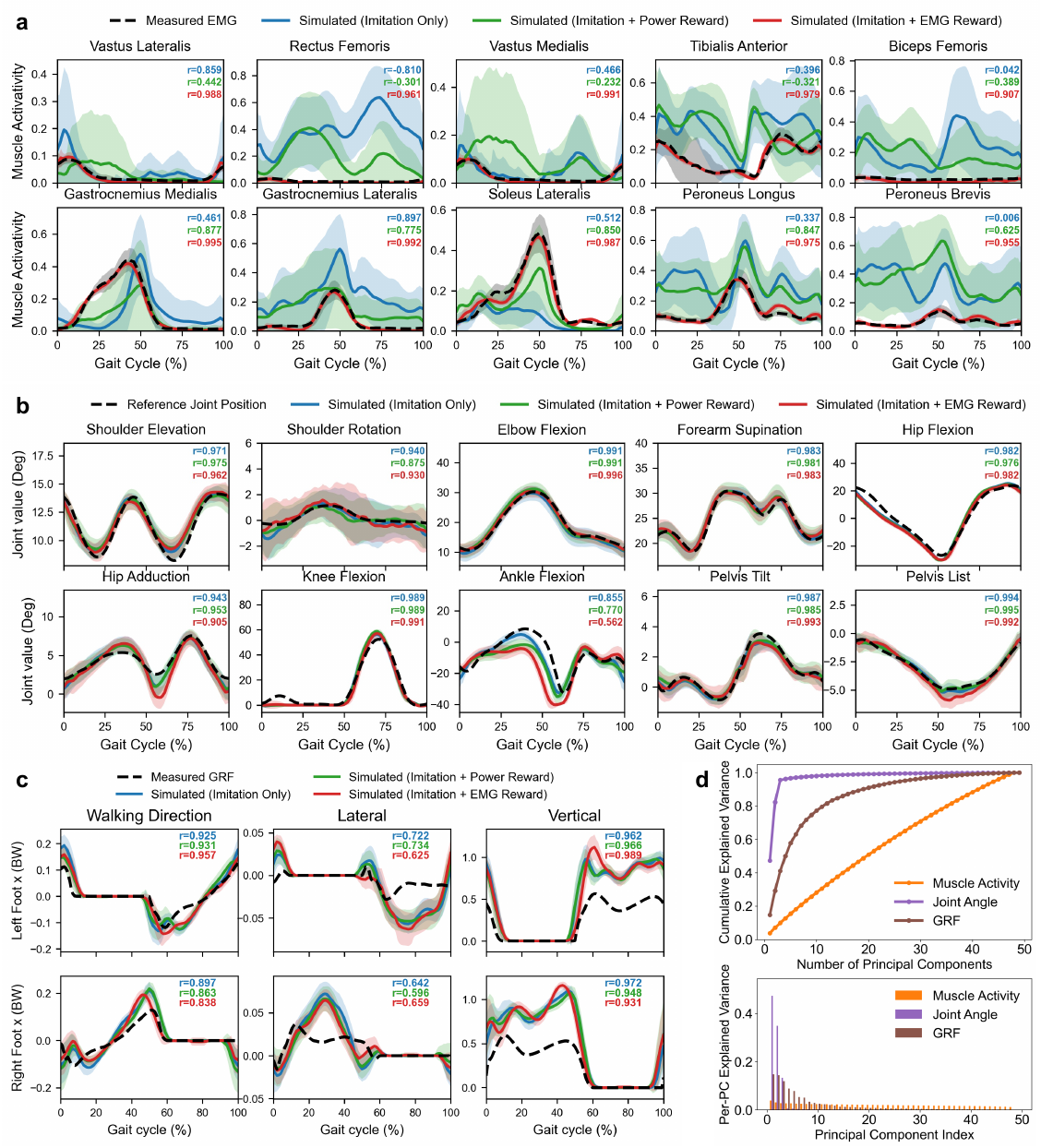}
\caption{\textbf{Analysis of musculoskeletal dynamics during walking.} \textbf{a,} Simulated muscle activity compared with measured EMG over a single gait cycle. Lines denote the mean and shaded regions denote one standard deviation. \textbf{b,} Simulated and reference joint kinematics over a single gait cycle. \textbf{c,} Simulated and measured ground reaction forces for the left and right feet over a single gait cycle, with the value normalized by body weight (BW). \textbf{d,} Principal component analysis of simulated full-body muscle-activity, joint-angle and ground-reaction-force time series from 50 trained policies, showing cumulative and per-component explained variance.
In \textbf{a-c}, lines denote the mean and shaded regions denote one standard deviation. Text indicates the Pearson correlation coefficient between simulated and measured signals.}
\label{fig_5}
\end{figure}

Distinct muscle-control solutions emerged across these training conditions (Fig.~\ref{fig_5}a). As expected, policies trained with EMG regularization showed the closest correspondence to the measured EMG across all ten recorded muscles, with a mean Pearson correlation of $r=0.973$ (range, $0.907$--$0.995$), substantially exceeding both imitation-only policies ($r=0.479$; range, $0.006$--$0.897$) and power-regularized policies ($r=0.566$; range, $0.232$--$0.877$). This EMG agreement reflects optimization under an auxiliary constraint rather than independent physiological validation. It nevertheless shows that the framework can steer learning towards distinct, EMG-consistent regions of the redundant muscle-actuation space without sacrificing accurate motion reproduction. Power regularization likewise shifted the learned solutions relative to imitation alone for several muscles, consistent with the emergence of a more economical walking strategy \cite{zarrugh1974optimization,kuo2005energetic}. Together, these results indicate that \framework~can uncover and selectively navigate a broad solution space of musculoskeletal control by learning directly in the native muscle-actuation space.

Despite these pronounced differences in muscle activation, policies trained under different regularization schemes produced highly similar joint trajectories over the gait cycle and closely matched the reference motion (Fig.~\ref{fig_5}b). The same convergence was observed at the level of external mechanics: simulated ground-reaction forces reproduced the measured temporal profiles of both feet with consistently high correlations across methods (mean Pearson correlation $r=0.853$ for imitation-only, $0.833$ for EMG regularization, and $0.840$ for power regularization; Fig.~\ref{fig_5}c). Agreement was strongest in the vertical component, for which the correlation exceeded $0.93$ under all training settings. These analyses show that multiple internal muscle-actuation strategies can support nearly identical kinematic and external biomechanical behavior.

To quantify this dissociation between externally observable motion and internally generated musculoskeletal control, we performed principal component analysis on simulated 700-muscle activity, joint-angle and ground-reaction-force time series from all trained policies. Joint kinematics occupied the most compact low-dimensional subspace, with the first principal component alone explaining roughly half of the total variance. Ground-reaction forces also showed low complexity, with cumulative variance saturating within only a few components. By contrast, muscle activity was distributed across a substantially broader space, with variance accumulating gradually over many components (Fig.~\ref{fig_5}d). Together, these results show that near-identical walking kinematics can be supported by diverse internal musculoskeletal dynamics, highlighting the substantial redundancy and flexibility of full-body human motor control.

\section*{Discussion}\label{Discussion}

The study of neuro-actuated musculoskeletal dynamics during human motion has long been constrained by severe computational bottlenecks and algorithmic inefficiency. Here we present a scalable musculoskeletal behavior emulation framework that overcomes these barriers. By executing reinforcement learning entirely within a GPU-native physics pipeline, the framework shifts the emphasis from reproducing isolated movement instances to enabling robust, large-scale statistical analysis. This architecture makes complex whole-body musculoskeletal modeling computationally practical and supports efficient, high-fidelity motion tracking on a single consumer-grade GPU. More broadly, it establishes a computational foundation for analyzing internal musculoskeletal dynamics and for studying how specificity and diversity are jointly expressed in human embodied intelligence.

Specificity and diversity are both essential for neuromusculoskeletal behavior. Specificity is required for precise, targeted reproduction of a desired movement. Diversity is required because the redundant musculoskeletal system admits many feasible internal control strategies for the same external task. In a high-dimensional musculoskeletal system, however, standard deep reinforcement learning agents often suffer from vanishing exploration. We therefore introduce value-guided flow exploration, which goes beyond hardware acceleration alone to mitigate the curse of dimensionality while preserving the full flexibility of the native actuation space. Conventional motion-tracking frameworks also depend heavily on manually engineered reward weights, which can introduce human bias and limit robust generalization across behaviors. Our framework instead employs an adversarial discriminator to adaptively balance tracking errors without manual parameter tuning. Together, these algorithmic advances enable rapid convergence on complex tracking tasks and support high-fidelity kinematic reproduction across versatile, agile and highly dynamic motions.

This efficiency substantially expands our ability to study neuro-actuated musculoskeletal dynamics during full-body motion. We trained ensembles of expert walking policies under different regularization schemes, thereby probing the high-dimensional null space of the severely underdetermined inverse-dynamics problem. Distinct and widely distributed muscle-activation strategies converged to nearly identical and highly accurate whole-body kinematics. This result provides a large-scale computational illustration of Bernstein's redundancy problem \cite{bernshtein1967co}, showing that the musculoskeletal plant itself affords a broad manifold of feasible solutions. In this sense, specificity and diversity emerge as complementary rather than opposing properties: the system can preserve highly specific task-relevant external behavior while sustaining diverse internal control solutions. Although the current framework optimizes distinct policies rather than explicitly modeling trial-to-trial biological motor noise, the emergence of this broad and flexible solution space is conceptually consistent with theories of biological coordination such as the uncontrolled manifold hypothesis \cite{scholz1999uncontrolled}. More broadly, these results highlight how an over-actuated musculoskeletal system can support substantial internal variability in redundant dimensions while preserving task-relevant external kinematics and mechanics, thereby reconciling diversity of internal control with specificity of observable behavior.

Although we demonstrate the framework on a complex whole-body human musculoskeletal model, its overall design should in principle be extendable to other species with appropriate replacement and reparameterization of the underlying biomechanical model. \framework~could therefore be adapted to simulate and investigate the musculoskeletal dynamics of diverse organisms, including fruit flies \cite{ozdil2025musculoskeletal}, mice \cite{dewolf2024neuro}, and other species \cite{stark2021three, la2021ostrichrl, van2024muscle}. This generality further underscores the broader relevance of specificity and diversity as organizing principles across biological and computational systems.

With these advances, one may tackle more challenges in the future. Our current simulations assume idealized instantaneous sensory feedback and therefore omit signal-dependent noise and transmission delays, both are fundamental to biological motor control \cite{harris1998signal, franklin2011computational}. The Hill-type muscle models used here are simplified and do not fully capture key features such as history-dependent force production and fatigue \cite{millard2013flexing, feng2023musclevae}. Although EMG regularization improves alignment with measured muscle activity, it can reduce tracking accuracy in mechanically related joints, such as the ankle, underscoring the fact that surface EMG does not map one-to-one onto muscle-tendon dynamics \cite{disselhorst2009surface}. The rigid-body formulation also limits the fidelity of contact mechanics, particularly for quantities such as ground-reaction forces, motivating future integration with more advanced soft-body contact models \cite{gong2025contact}. Finally, although repeated training trials introduce diversity through stochastic policy initialization, the current framework relies on a single generic human morphology and does not yet capture the anatomical and physiological variability observed across human populations.

By transforming musculoskeletal motion imitation into an efficient, high-throughput computational pipeline, our work provides a powerful platform for systematic emulation and analysis of the specificity and diversity of complex neuro-actuated musculoskeletal dynamics.

\section*{Methods}\label{Methods}

\subsection*{Whole-body musculoskeletal model}

To enable high-fidelity simulation of human movement, we use a version of the open-sourced MS-Human-700 whole-body musculoskeletal model \cite{zuo2024self}. 
The model is actuated by 700 individual muscle-tendon units (MTUs), providing a highly redundant actuation system characteristic of biological organisms.

\noindent \textbf{Muscle-tendon dynamics.} The force-generation properties of muscles were modeled using a Hill-type muscle model \cite{zajac1989muscle}. The muscle force $F_m$ is determined by the muscle activation $act$, normalized fiber length $l_m$, and normalized fiber contraction velocity $v_m$, governed by the following equation:
\begin{equation}
    F_m(\text{act}, l_m, v_m) = F_{\text{max}} \cdot \left[ \text{act} \cdot f_l(l_m) \cdot f_v(v_m) + f_p(l_m) \right]
\label{flv}
\end{equation}
where $F_{\text{max}}$ is the maximum isometric force of the muscle. The function $f_l(l_m)$ represents the active force-length relationship, capturing the dependence of force production on fiber length. The function $f_v(v_m)$ describes the force-velocity relationship, reflecting the reduced force capacity during concentric contraction and enhanced force during eccentric contraction. The passive force-length function $f_p(l_m)$ models the elastic resistance of stretched muscle fibers. These force-length and force-velocity curves were parameterized based on experimental data from biomechanical studies \cite{millard2013flexing}.

Muscle activation dynamics were modeled as a first-order process to capture the electromechanical delay between neural excitation and force production:
\begin{equation}
    \frac{d\text{act}}{dt} = \frac{u - \text{act}}{\tau(u, \text{act})}, \quad
    \tau(u, \text{act}) = \begin{cases}
        \tau_{\text{act}} (0.5 + 1.5\text{act}) & u > \text{act} \\
        \tau_{\text{deact}} / (0.5 + 1.5\text{act}) & u \le \text{act}
    \end{cases}
\label{excitation-activation}
\end{equation}
where $u \in [0, 1]$ is the neural excitation signal to the muscle, and $\tau_{\text{act}}$ and $\tau_{\text{deact}}$ are the activation and deactivation time constants, set to 10 ms and 40 ms, respectively \cite{winters1995improved}. This nonlinear formulation captures the asymmetry between rapid muscle activation and slower relaxation observed in physiological measurements. 

\noindent \textbf{Musculoskeletal dynamics.} The full-body dynamics were simulated by solving the equations of motion in generalized coordinates using the Euler-Lagrangian formulation:
\begin{equation}
    M(q)\ddot{q} + C(q, \dot{q}) = J_m(q)^T F_{\text{mtu}} + \tau_{\text{ext}}
\end{equation}
where $q, \dot{q}, \ddot{q}$ are the joint position, velocity, and acceleration vectors in generalized coordinates. $M(q)$ is the generalized mass matrix encoding the inertial properties of the body segments. $C(q, \dot{q})$ captures Coriolis, centrifugal, and gravitational forces. $\tau_{\text{ext}}$ represents external forces arising from contact in physical simulation. $J_m(q)$ is the muscle moment arm matrix that maps muscle forces $F_{\text{mtu}}$ to joint torques, accounting for the geometric routing of muscle-tendon paths across joints \cite{uchida2021biomechanics}.

\noindent \textbf{Contact modeling.} Ground contact was modeled using the soft contact model implemented in MuJoCo\cite{todorov2012mujoco}. Contact geometry was defined for the feet (heel, metatarsal, and toe regions) and hands (metacarpal and fingers), enabling realistic ground-reaction forces during movement.

\subsection*{GPU-accelerated simulation framework}

\noindent \textbf{System architecture.} Our framework integrates the MS-Human-700 model with MuJoCo Warp, a GPU-native simulation backend implemented with NVIDIA Warp kernels. By locating physics simulation and neural network inference both on the GPU, the system avoids the repeated host-device transfers that commonly limit the efficiency of hybrid CPU simulation and GPU learning pipelines, thereby enabling higher training throughput. Compared with conventional humanoid and robotics simulation, whole-body musculoskeletal dynamics induces a larger and more variable constraint set due to anatomical joint coupling (e.g., scapulohumeral rhythm within the shoulder complex) and richly distributed contact across the body. Such variability is challenging for compilation-based accelerators such as MuJoCo XLA (MJX), whose execution is most efficient under static array shapes. In our setting, however, the number of active contacts and constraints changes substantially across timesteps. MuJoCo Warp is advantageous in this regime because it naturally supports dynamically varying active constraint sets, avoiding the inefficiencies associated with fixed-shape allocation in highly constraint-variable musculoskeletal simulation.

\noindent \textbf{Simulation parameters.} The physics simulation operated at a timestep of $\Delta t_{\text{sim}} = 0.002$ s to ensure numerical stability for the muscle dynamics and contact resolution. The control policy was queried at a lower frequency with a control timestep of $\Delta t_{\text{ctrl}} = 0.02$ s, corresponding to a control frequency of 50~Hz. This resulted in 10 simulation substeps per control step, balancing computational efficiency with simulation fidelity. The parameters of human musculoskeletal model (e.g., muscle activation-deactivation time, muscle force and joint damping) are consistent in all motion tracking tasks.


\subsection*{Reference motion data and kinematic retargeting}

\noindent \textbf{Motion capture datasets.} Reference motion trajectories in Fig.~\ref{fig_3} were derived from the AMASS dataset \cite{mahmood2019amass}, a large-scale collection of human motion capture data unified under the SMPL-X body representation \cite{pavlakos2019expressive}. The dataset encompasses diverse motor behaviors including locomotion (walking, running), dynamic movements (jumping, kicking), and skilled actions (dancing, cartwheeling). For the solution space exploration experiments, we used walking data from the Gait 120 dataset \cite{boo2025comprehensive}, which provides synchronized motion capture, ground reaction force (GRF) and surface electromyography (sEMG) recordings from 120 adult male subjects.

Physics-based motion imitation with musculoskeletal systems requires high-fidelity reference trajectories. Artifacts such as ground penetration, foot sliding, and self-intersections in the reference motion can severely degrade reinforcement learning performance, as the policy must expend capacity compensating for physically inconsistent targets \cite{araujo2025retargeting}. We therefore developed a retargeting pipeline to accurately map SMPL-X motion capture data to the joint space of the musculoskeletal model.

\noindent \textbf{Body shape calibration.} To retarget motion capture data collected on varying body morphologies to the generic MS-Human-700 model, we first established a correspondence between the two body representations. We labeled a set of osseous marker sites on the MS-Human-700 model corresponding to anatomical landmarks commonly used in motion capture experiments (e.g., anterior superior iliac spine) \cite{keller2023skin}. We then identified the corresponding vertex indices on the SMPL-X mesh by matching anatomical positions. Using these marker correspondences in the T-pose, we iteratively optimized the pose parameters $\theta\in\mathbb{R}^{21\times3}$ and shape parameters $\beta\in\mathbb{R}^{300}$ as well as root orientation $\Gamma\in\mathbb{R}^3$ and translation $\tau\in\mathbb{R}^3$. By minimizing the Euclidean distance between two marker sets, a personalized shape $\beta^*$ that approximates the skeletal proportions of the musculoskeletal model was identified and served as a reliable and unified body shape initialization for inverse kinematics (Fig.~\ref{fig_1}b). 

\noindent \textbf{Marker-based inverse kinematics.} For each frame of the motion capture sequence, the target positions for inverse kinematics were obtained by reconstructing the SMPL-X mesh with the optimized shape $\beta^*$ and the pose parameters from the AMASS data. Additionally, we defined two sets of corresponding markers: joint markers positioned at skeletal landmarks mapped to SMPL-X joint locations, and skin markers positioned at surface anatomical landmarks mapped to SMPL-X mesh vertices.

The joint configuration $q^{\text{ref}}$ of the MS-Human-700 model was computed by solving an inverse kinematics optimization problem:
\begin{equation}
    q^{\text{ref}} = \arg\min_{q} \sum_{i \in \mathcal{M}_j} w_i^j \| p_i(q) - p_i^{\text{target}} \|^2 + \sum_{i \in \mathcal{M}_s} w_i^s \| p_i(q) - p_i^{\text{target}} \|^2 + \lambda_q \| q - q_{\text{prev}} \|^2
\end{equation}
where $\mathcal{M}_j$ and $\mathcal{M}_s$ denote the sets of joint and skin markers, respectively. $p_i(q)$ is the marker position computed via forward kinematics given joint configuration $q$, $p_i^{\text{target}}$ is the target position from the SMPL-X model, and $w_i^j$, $w_i^s$ are marker-specific weights. The regularization term penalizes deviation from the previous frame's configuration $q_{\text{prev}}$ to ensure temporal smoothness. We solved this optimization using Mink\cite{mink}, which formulates the problem as a quadratic program over generalized velocities. To prevent physically implausible solutions, we incorporated collision avoidance constraints between the contact geometries and the ground plane, ensuring that the retargeted motion maintained consistent contact states without ground penetration.

\noindent \textbf{Trajectory filtering and post-processing.} The raw inverse kinematics output was smoothed using a second-order Butterworth low-pass filter with a cutoff frequency of 20 Hz to remove high-frequency noise while preserving the essential motion dynamics. We then computed the global positions of key body segments using forward kinematics and calculated reference joint velocities $\dot{q}^{\text{ref}}$ using finite differences. The minimum foot height across the entire sequence was subtracted from the root height to avoid floating artifacts.

\noindent \textbf{Subject selection for validation.} For the solution space exploration experiments, we selected the subject from the Gait 120 dataset whose anthropometric measurements (height: 1.742 m, mass: 61.0 kg) closely matched those of the MS-Human-700 model (height: 1.735 m, mass: 60.0 kg). This subject's motion capture data served as the reference trajectory for policy training, while the synchronized GRF and sEMG recordings provided ground truth for validating simulated outputs. The Gait 120 dataset provides raw marker trajectories. We therefore manually calibrated the marker set on the MS-Human-700 model to match the anatomical landmarks used in the motion-capture protocol and then applied the same inverse-kinematics pipeline described above. The raw data consisted of single gait cycles per trial; for continuous training, we applied phase alignment, cycle averaging, and boundary smoothing to generate extended multi-cycle trajectories.

\subsection*{Motion imitation using deep reinforcement learning}

We formulated the motor control problem as a Markov Decision Process (MDP) defined by the tuple $(\mathcal{S}, \mathcal{A}, \mathcal{P}, \mathcal{R}, \gamma)$, where $\mathcal{S}$ is the state space, $\mathcal{A}$ is the action space, $\mathcal{P}$ is the transition dynamics, $\mathcal{R}$ is the reward function, and $\gamma$ is the discount factor.

\noindent \textbf{Observation space.} The state $s_t \in \mathcal{S}$ provided to the policy at each control step includes:
\begin{itemize}
    \item \texttt{Proprioceptive state}: Joint positions $q_t \in \mathbb{R}^{85}$ and velocities $\dot{q}_t \in \mathbb{R}^{85}$ in generalized coordinates. Key body segments position $\text{xpos}_{b,t} \in \mathbb{R}^{30}$, orientation $\text{xmat}_{b,t} \in \mathbb{R}^{60}$.
    \item \texttt{Muscle state}: Muscle activations $\text{act}_t \in \mathbb{R}^{700}$, muscle forces $\text{F}_{m,t} \in \mathbb{R}^{700}$, normalized muscle fiber lengths $l_{m,t} \in \mathbb{R}^{700}$, and fiber velocities $\dot{l}_{m,t} \in \mathbb{R}^{700}$.
    \item \texttt{Reference trajectory}: Target joint positions $q_t^{\text{ref}}$, key body positions $\text{xpos}_t^{\text{ref}}$ and key body orientations $\text{xmat}_t^{\text{ref}}$ from the retargeted motion capture data.
\end{itemize}

\noindent \textbf{Action space.} The action $a_t \in \mathcal{A} = [0, 1]^{700}$ consists of continuous neural excitation signals for all 700 muscles. These excitation signals are processed by the muscle activation dynamics (Eq. \ref{excitation-activation}) to produce muscle activations, which in turn generate muscle forces through the Hill-type model (Eq. \ref{flv}).

\noindent\textbf{Unified imitation environment.} We designed a unified imitation environment that transforms simulator states into observations and reward signals for motion tracking (Fig.~\ref{fig_2}b). Kinematic information and muscle states are extracted from the full simulation data to form the high-dimensional observation $\boldsymbol{s}$ along with reference motion. The discrepancies in body positions and joint angles between the simulated and reference motions are then computed as a raw tracking-error vector $\boldsymbol{\Delta}$. To convert this high-dimensional error into a scalar learning signal, we adopt an adversarial differential discriminator \cite{zhang2025physics}, which adaptively maps the tracking error to a reward by training a discriminator $D$ to distinguish $\boldsymbol{\Delta}$ from a zero vector:
\begin{align}
    \mathcal{L}_{D} = -\log D(\boldsymbol{0}) - \log\!\left(1-D(\boldsymbol{\Delta})\right) + \lambda \norm{\nabla D}^2,
\end{align}
where the final term is a gradient penalty that regularizes the discriminator and smooths its prediction. The imitation reward is given by:
\begin{align}
\label{eq:tracking_r}
    r = -\log(1-D(\boldsymbol{\Delta}))
\end{align}
This unified imitation environment dynamically aggregates high-dimensional tracking errors and provides adaptive supervision that depends on both the motion type and the stage of learning. To facilitate the learning of these complex and extended sequences, we employed reference state initialization \cite{peng2018deepmimic} and adaptive sampling \cite{liao2025beyondmimic}, allowing the policy to sample more frequently from kinematic phases with higher failure rates.

\noindent \textbf{Reinforcement learning with flow-based exploration.} We incorporate value-guided flow exploration \cite{wei2026scalable} into an on-policy RL pipeline, enabling scalable and directed exploration in high-dimensional action spaces (Fig.~\ref{fig_2}c).

We parameterize the policy by a Gaussian initial sampler $\pi^{(0)}(\boldsymbol{a}\mid\boldsymbol{s})$ and a flow velocity field $\psi(\boldsymbol{a}\mid t,\boldsymbol{s},\boldsymbol{a}^{(t)})$. Given an observation $\boldsymbol{s}$, a base action is first sampled from $\pi^{(0)}$, and is then progressively refined by $\psi$ into the final action:
\begin{align}
    \label{eq:pi_ode}
        \boldsymbol{a} = \boldsymbol{a}^{(1)} = \boldsymbol{a}^{(0)} + \int_0^1 \psi(t,\boldsymbol{s},\boldsymbol{a}^{(t)})\,dt,
        \quad
    \boldsymbol{a}^{(0)} \sim \pi^{(0)}(\cdot\mid\boldsymbol{s}).
\end{align}
This flow-based parameterization enables expressive action sampling beyond isotropic Gaussian policies, allowing the policy to better approximate complex target distributions.

Using the replayed trajectory buffer $\mathcal{B}=\{(\boldsymbol{s},\boldsymbol{a},r,\boldsymbol{s}')\}$, we train the policy towards a target distribution specified by a value-guided probability flow:
\begin{align}
    \frac{d\boldsymbol{a}^{(t)}}{dt} = \psi^{(t)}_Q( \boldsymbol{a}^{(t)}; \boldsymbol{s}) = \nabla_{\boldsymbol{a}^{(t)}} Q(\boldsymbol{s}, \boldsymbol{a}^{(t)}), \quad \boldsymbol{a}^{(0)}\sim \pi^{(0)}(\cdot|\boldsymbol{s}),
\end{align}
where $Q(\boldsymbol{s},\boldsymbol{a})$ denotes the learned state-action value function. Sampling along this value-guided flow yields higher-value actions with policy-improvement guarantees. Because $Q$ is not explicitly learned in standard on-policy RL, we approximate it using the learned state value function $V(\boldsymbol{s})$ and the estimated advantage function $A(\boldsymbol{s},\boldsymbol{a})$:
\begin{align}
    \mathcal{L}_Q = \mathbb{E}_{(\boldsymbol{o, a}) \sim \mathcal{B}}\norm{Q(\boldsymbol{s}, \boldsymbol{a})-V(\boldsymbol{s}) - A(\boldsymbol{s}, \boldsymbol{a})}^2.
\end{align}
This formulation avoids additional bootstrapping and therefore preserves the training stability of on-policy RL.

During policy optimization, the initial sampler $\pi^{(0)}$ is updated by policy gradient, together with ratio clipping and entropy regularization for stabilization \cite{schulman2017proximal}. The flow velocity field $\psi$ is trained to approximate the value-guided probability flow:
\begin{align}
    \mathcal{L}_{\psi} = \norm{\psi(t, \boldsymbol{s}, \boldsymbol{a}^{(t)})-\psi^{(t)}_Q( \boldsymbol{a}^{(t)}; \boldsymbol{s})}^2.
\end{align}
The objective $\mathcal{L}_{\psi}$ can be efficiently optimized by flow matching \cite{lipman2022flow}. Overall, the flow-based policy promotes effective exploration towards high-value regions of high-dimensional action spaces, substantially improving the learning efficiency of musculoskeletal motion control.

\noindent \textbf{Training hyperparameters.} The policy was optimized using PPO with the following hyperparameters: rollout steps $h=8$, discount factor $\gamma = 0.99$, GAE parameter $\lambda = 0.95$, clip ratio $\epsilon = 0.3$, learning rate $\alpha = 3 \times 10^{-5}$, entropy coefficient $0.0005$, and 5 epochs per batch update. For the training of flow velocity field, we adopted the default hyperparameters in the original paper, using $N=20$ gradient steps with step size $\eta=0.01$ to construct the target probability flow, and employ $N_{\text{ODE}}=20$ steps with $\Delta t_{\text{ODE}}=0.05$ when refining the base action. Each training trail used 2,048 parallel environments. For all the neural network components (initial sampler, flow velocity field, value function, discriminator), we employ 3-layer MLP with 1024 hidden units.





\subsection*{Evaluation metrics}

\noindent \textbf{Kinematic tracking accuracy.} Motion tracking performance was quantified using mean joint angle error (in degrees) averaged across all DOFs and time steps:
\begin{equation}
    E_{\text{joint}} = \frac{1}{T \cdot J} \sum_{t=1}^{T} \sum_{d=1}^{J} | q_t^d - q_t^{\text{ref},d} |
\end{equation}
where $J = 79$ is the number of joints excluding root and $T$ is the trajectory length.

Body segments tracking accuracy was computed as the Euclidean distance error between simulated and reference positions of the bodies:
\begin{equation}
    E_{\text{body}} = \frac{1}{T \cdot N_{\text{body}}} \sum_{t=1}^{T} \sum_{i=1}^{N_{\text{body}}} \| x_{b,t}^i - x_{b,t}^{\text{ref},i} \|
\end{equation}


\noindent \textbf{Simulation throughput.} Computational performance was measured as steps per second (SPS), defined as the total number of environment steps completed per second of wall-clock time across all parallel environments. For the throughput benchmark in Fig.~\ref{fig_4}a, all GPU measurements were obtained on NVIDIA GeForce RTX 5090 GPUs. CPU baselines were evaluated on a 4-socket server equipped with Intel(R) Xeon(R) Platinum 8180 CPUs @ 2.50\,GHz, with 224 CPU cores in total (4 physical CPUs $\times$ 56 cores), and 1 NVIDIA GeForce RTX 4070 Ti SUPER GPU. The CPU training baseline used the conventional Stable-Baselines3 pipeline.

\subsection*{Analysis of neuro-actuated musculoskeletal dynamics}\label{method:analysis}

To investigate neuro-actuated redundancy and probe feasible control solutions, we conducted a large-scale statistical analysis using our framework.

\noindent \textbf{Training regularization.}
We considered three training schemes in this analysis by modifying the reward function in the motion-imitation environment.

\begin{itemize}
    \item \texttt{Imitation Only}: 
Training was performed using only the adaptive tracking reward $r_{\text{tracking}}$ defined in Eq.~(\ref{eq:tracking_r}).
    \item \texttt{Imitation + EMG Reward}:
To encourage agreement between simulated muscle activity and experimentally measured EMG during motion tracking, we added an EMG-matching reward:
\begin{align}
    r = r_{\text{tracking}} + w_{\text{EMG}} r_{\text{EMG}}, \quad r_{\text{EMG}} = -\frac{1}{n_{\text{channel}}} \sum_{i}^{n_{\text{channel}}} \left(\text{act}^{(i)}_{\text{measure}}-\text{act}^{(i)}_{\text{sim}}\right)^2,
\end{align}
where $n_{\text{channel}} = 10$ is the number of EMG channels, $\text{act}_{\text{measure}}^{(i)}$ denotes the measured EMG signal from the $i$th channel, and $\text{act}_{\text{sim}}^{(i)}$ denotes the simulated activity of the corresponding muscle. We set the EMG reward weight to $w_{\text{EMG}} = 100$ to match the scale of the tracking reward.
    \item \texttt{Imitation + Power Reward}: 
To encourage energetically economical control during motion tracking, we added a power-penalty reward:
\begin{align}
    r = r_{\text{tracking}} + w_{\text{power}} r_{\text{power}}, \quad r_{\text{power}} = -\frac{1}{|\mathcal{A}|} \sum_{i}^{|\mathcal{A}|} \left|F_m^{(i)}\cdot v_m^{(i)}\right|,
\end{align}
where $|\mathcal{A}| = 700$ is the total number of muscles, and $F_m^{(i)}$ and $v_m^{(i)}$ denote the force and contraction velocity of the $i$th muscle, respectively. We varied the power reward weight as $w_{\text{power}} \in \{0.01, 0.05, 0.1\}$ to train an ensemble of policies.
\end{itemize}




\noindent \textbf{Biomechanical comparison.} 
To assess the physiological plausibility of the simulated muscle-activation patterns, we compared the simulated results with surface electromyography (sEMG), joint-angle and ground-reaction-force (GRF) data from the selected subject performing level walking in the Gait-120 dataset. All reference data were preprocessed by the dataset authors.

We generated simulated kinematic and biomechanical signals by rolling out the trained policies and recording muscle activations, joint angles, as well as GRFs between the feet and the ground. The raw simulated GRF signals were smoothed using a sliding-window average with a window size of 0.1\,s.

For the comparisons shown in Fig.~\ref{fig_5}a-c, we segmented both simulated and measured time series into individual walking cycles. For each signal, we computed the mean and standard deviation over a normalized gait cycle, as well as the Pearson correlation between simulated and measured signals.

\noindent \textbf{Analysis of internal musculoskeletal dynamics.} For the analysis of internal musculoskeletal dynamics shown in Fig.~\ref{fig_5}d, we flattened each simulated time series into a single vector, stacked the resulting vectors across 50 trained policies, and then performed principal component analysis separately for muscle activity, joint-angle trajectories, and ground-reaction-force trajectories.






\newpage
\renewcommand\bibnumfmt[1]{#1.}

\bibliography{sn-bibliography}

\end{document}